%% file: main.tex
\documentclass[10pt,twocolumn,letterpaper]{article}

\usepackage{cvpr}              


\definecolor{cvprblue}{rgb}{0.21,0.49,0.74}
\usepackage[pagebackref,breaklinks,colorlinks,allcolors=cvprblue]{hyperref}
\usepackage{multirow}
\usepackage{adjustbox}
\usepackage{colortbl}
\usepackage{arydshln}
\usepackage{tikz}
\usepackage{tikz-3dplot}
\usetikzlibrary{bending, external, plotmarks,shapes,positioning}
\usepackage{lipsum}

\pgfdeclarelayer{background}
\pgfdeclarelayer{foreground}
\pgfsetlayers{background,main,foreground}

\definecolor{matlab1}{rgb}{0.00000,0.44700,0.74100}%
\definecolor{matlab2}{rgb}{0.85000,0.32500,0.09800}%
\definecolor{matlab3}{rgb}{0.92900,0.69400,0.12500}%
\definecolor{matlab4}{rgb}{0.49400,0.18400,0.55600}%
\definecolor{matlab5}{rgb}{0.4660, 0.6740, 0.1880}%
\definecolor{matlab6}{rgb}{0.3010, 0.7450, 0.9330}%
\definecolor{matlab7}{rgb}{1.0, 0.0, 0.0}%

\def\paperID{7633} 
\def\confName{CVPR}
\def\confYear{2025}

\definecolor{somegray}{rgb}{0.5, 0.5, 0.5}
\newcommand{\darkgrayed}[1]{\textcolor{somegray}{#1}}
\makeatletter
\newcommand*\titleheader[1]{\gdef\@titleheader{#1}}
\AtBeginDocument{%
  \let\st@red@title\@title
  \def\@title{%
    \vskip-3em
    \bgroup\normalfont\large\centering\@titleheader\par\egroup
    \vskip1.5em\st@red@title}
}
\makeatother

\titleheader{\darkgrayed{This paper has been accepted for publication at the\\
IEEE Conference on Computer Vision and Pattern Recognition (CVPR), Nashville, 2025.
\copyright IEEE}}

\title{GG-SSMs: Graph-Generating State Space Models}

\author{Nikola Zubi\'{c} and Davide Scaramuzza \\
Robotics and Perception Group, University of Zurich, Switzerland\\ \\
}

\begin{document}
\maketitle
\input{sec/0_abstract}    
\input{sec/1_intro}
\input{sec/2_related_work}
\input{sec/3_method}
\input{sec/4_experiments}
\input{sec/5_ablation_study}
\input{sec/6_conclusion}
\input{sec/7_acknowledgment}

\input{sec/X_suppl}

{
    \small
    \bibliographystyle{ieeenat_fullname}
    \bibliography{main}
}

\end{document}

%% file: sec/0_abstract.tex
\begin{abstract} 
State Space Models (SSMs) are powerful tools for modeling sequential data in computer vision and time series analysis domains. However, traditional SSMs are limited by fixed, one-dimensional sequential processing, which restricts their ability to model non-local interactions in high-dimensional data. While methods like Mamba and VMamba introduce selective and flexible scanning strategies, they rely on predetermined paths, which fails to efficiently capture complex dependencies.
We introduce Graph-Generating State Space Models (GG-SSMs), a novel framework that overcomes these limitations by dynamically constructing graphs based on feature relationships. Using Chazelle's Minimum Spanning Tree algorithm, GG-SSMs adapt to the inherent data structure, enabling robust feature propagation across dynamically generated graphs and efficiently modeling complex dependencies.
We validate GG-SSMs on 11 diverse datasets, including event-based eye-tracking, ImageNet classification, optical flow estimation, and six time series datasets. GG-SSMs achieve state-of-the-art performance across all tasks, surpassing existing methods by significant margins. Specifically, GG-SSM attains a top-1 accuracy of 84.9\% on ImageNet, outperforming prior SSMs by 1\%, reducing the KITTI-15 error rate to 2.77\%, and improving eye-tracking detection rates by up to 0.33\% with fewer parameters.
These results demonstrate that dynamic scanning based on feature relationships significantly improves SSMs' representational power and efficiency, offering a versatile tool for various applications in computer vision and beyond.
\end{abstract}

%% file: sec/1_intro.tex
\hfill \break
\noindent \textbf{Multimedial Material:} For GitHub code, poster and other details visit \url{https://github.com/uzh-rpg/gg_ssms}. 

\begin{figure}[t]
\centering
\includegraphics[width=0.8\linewidth]{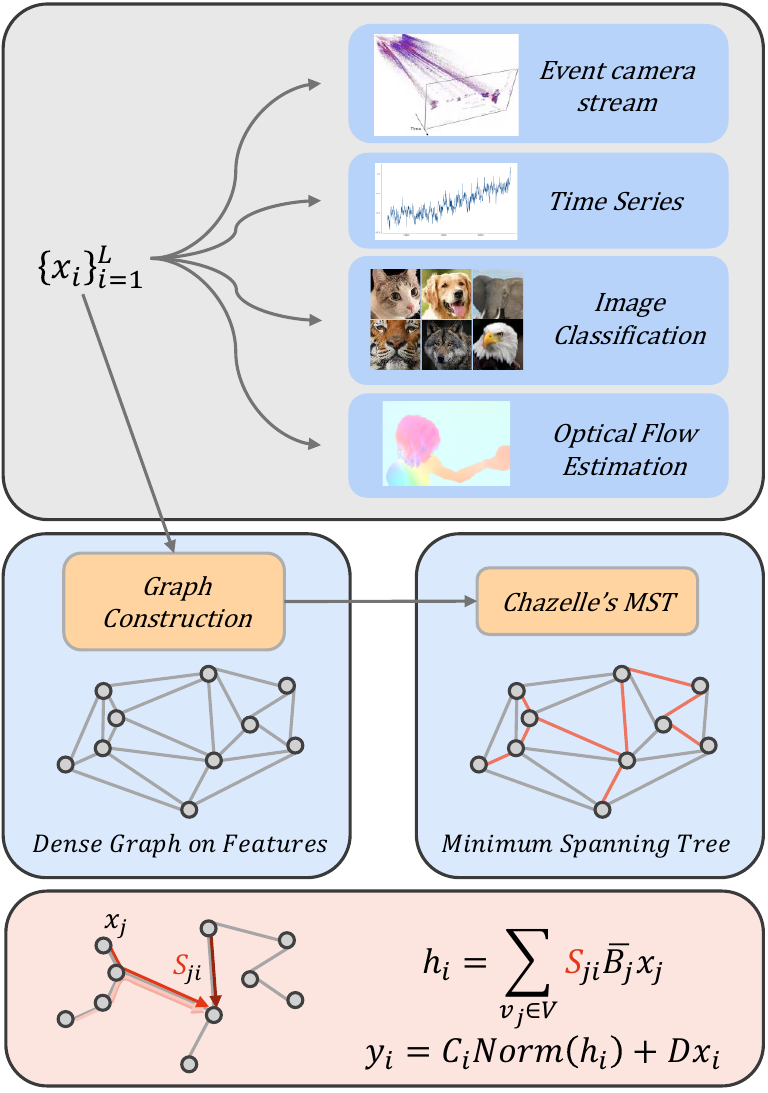}
\caption{\textbf{Illustration of the Graph-Generating State Space Model (GG-SSM).} Given an input feature set $\{\mathbf{x}_i\}_{i=1}^L$, we construct a graph based on feature dissimilarities and apply an efficient algorithm to generate a minimum spanning tree $\mathcal{T}$. SSM state propagation is then performed along this tree to obtain improved feature representations.}
\label{fig:gg_ssm_core}
\end{figure}

\section{Introduction}
\label{sec:introduction}

Modeling complex, long-range dependencies in high-dimensional data is a fundamental challenge in computer vision and time series analysis. Accurately capturing these dependencies is crucial for understanding intricate structures and relationships within data, directly impacting the performance of machine learning models in real-world applications.

Traditional models like Convolutional Neural Networks (CNNs) \cite{he2016deep} excel at capturing local patterns but struggle with global context due to their limited receptive fields. Transformers \cite{vaswani2017attention} address this limitation by employing self-attention mechanisms to model global relationships; however, they suffer from quadratic computational complexity with respect to input size, making them less practical for high-resolution images or long sequences.

State Space Models (SSMs) have emerged as efficient alternatives for sequential data modeling, demonstrating the ability to capture long-range dependencies with linear computational complexity concerning sequence length \cite{gu2021combining, gu2022efficiently}. SSMs process sequences using state transitions, making them suitable for tasks requiring memory of previous inputs. However, traditional SSMs are constrained by fixed, one-dimensional sequential processing, which limits their ability to model non-local interactions inherent in high-dimensional visual data \cite{nguyen2022s4nd}. This limitation hampers their effectiveness in capturing complex spatial dependencies critical for comprehensive visual understanding.

Recent methods like Mamba \cite{mamba} and VMamba \cite{liu2024vmamba} have attempted to overcome these constraints by introducing selective and flexible scanning strategies. Mamba \cite{mamba} introduces a selective mechanism to improve context awareness, while VMamba \cite{liu2024vmamba} extends this approach with more scanning paths to achieve global context. Despite these advancements, they rely on predetermined 1D scanning trajectories unrolled in a few directions (e.g., on an image grid) and struggle to adapt to the diverse and complex structures found in data. Consequently, \textit{they fail to efficiently capture intricate dependencies and long-range interactions not aligned with the predefined paths.}

The core challenge remains: designing models that can adaptively and efficiently capture complex, non-local dependencies in high-dimensional data without incurring prohibitive computational costs. Fixed scanning strategies cannot accommodate the diverse structures in data. While graph-based models naturally represent complex relationships \cite{kipf2017semi}, they often face high computational complexity when constructing and processing large graphs.

This paper introduces \textbf{Graph-Generating State Space Models (GG-SSMs)}, a novel framework that overcomes these limitations by dynamically constructing graphs based on feature relationships within the data. By utilizing Chazelle's Minimum Spanning Tree (MST) algorithm \cite{chazelle2000minimum}, which operates with near-linear time complexity, GG-SSMs adapt to the inherent data structure, enabling robust feature propagation across dynamically generated graphs. This approach efficiently models complex, long-range dependencies by moving beyond fixed scanning paths and capturing intrinsic relationships between features.

The key idea behind GG-SSMs is integrating dynamic graph construction into the state space modeling framework, allowing the model to naturally adapt its processing pathways based on the data's structure. GG-SSMs efficiently identify the most significant connections among data elements by constructing an MST over the data features. This dynamic graph is the backbone for state propagation, enabling the model to capture long-range dependencies and complex interactions without significant computational overhead.

Our contributions can be summarized as follows:

1. \textbf{Dynamic Graph-Based State Space Modeling}: We propose a method for integrating dynamic graph structures into SSMs, enabling the capture of complex spatial and temporal dependencies in high-dimensional data.

2. \textbf{Efficient Computation with MSTs}: Leveraging Chazelle's MST algorithm \cite{chazelle2000minimum}, GG-SSMs construct optimal graphs with minimal computational overhead, ensuring scalability to large datasets and high-resolution inputs.

3. \textbf{State-of-the-Art Performance}: GG-SSMs consistently outperform existing methods, including Mamba \cite{mamba} and VMamba \cite{liu2024vmamba}, across multiple benchmarks, achieving higher accuracy with fewer parameters and lower computational costs.

We validate GG-SSMs on 11 diverse datasets, covering various computer vision and time series tasks. Specifically:

- On the \textbf{ImageNet} classification benchmark \cite{Deng2009ImageNetAL}, GG-SSM achieves a top-1 accuracy of \textbf{84.9\%}, outperforming prior SSM-based models by \textbf{1\%}.

- In \textbf{optical flow estimation} on the KITTI-15 dataset \cite{geiger2012we}, GG-SSM reduces the error rate to \textbf{2.77\%}, the lowest reported to date.

- For \textbf{event-based eye-tracking} datasets (INI-30 \cite{Bonazzi2023RetinaL} and Event-based LPW \cite{Chen20233ETEE}), GG-SSM improves detection rates by up to \textbf{0.33\%} while using fewer parameters.

- Across six renowned \textbf{time series datasets}, GG-SSMs achieve superior forecasting accuracy, demonstrating versatility beyond traditional vision tasks.

These results demonstrate that dynamic scanning based on feature relationships significantly improves SSMs' representational power and efficiency, offering a versatile tool for various applications in computer vision and beyond.

%% file: sec/2_related_work.tex
\section{Related Work}
\label{sec:related_work}
\subsection{Challenges with Sequential-Only SSMs}
Data naturally exhibit multi-dimensional structures with complex spatial or spatiotemporal relationships in many computer vision applications and event-based data processing \cite{Zubic_2023_ICCV, Zubic2021}. Traditional State Space Models (SSMs) process signals in a strictly one-dimensional sequence, which makes them ill-suited to capture non-local interactions. For instance, images and event streams include spatial dependencies that span two or more dimensions, but standard SSMs restrict the flow of information along predetermined scanning paths.
While these 1D-sequential SSMs excel in tasks such as NLP, their inability to adapt to more complex data topologies leads to poor performance on high-dimensional or multi-variate inputs. Such limitations are particularly pronounced in tasks that demand a global understanding of spatial patterns, such as image classification, optical flow, and dynamic vision sensing.

\subsection{Approaches for High-Dimensional Visual SSMs}
Researchers have proposed specialized architectures for multi-dimensional data to address the shortcomings of sequential-only processing. S4ND~\cite{nguyen2022s4nd} pioneered a multi-dimensional approach by stacking independent 1D SSM modules, enabling limited handling of spatial axes. Baron et al.~\cite{Baron2024A2S} introduced a discrete multi-axial framework (A2S) incorporating a 2D-SSM spatial layer to better capture spatial dependencies. More recent advances explore sophisticated scanning and correlation techniques to improve spatial consistency, such as bidirectional~\cite{vim}, four-way~\cite{liu2024vmamba}, continuous~\cite{yang2024plainmamba}, zigzag~\cite{hu2024zigma}, window-based~\cite{huang2024localmamba}, and topology-based~\cite{oshima2024ssmmeetsvideodiffusion, Xiao2024GrootVLTT} processing.
Among these, Mamba~\cite{mamba} and VMamba~\cite{liu2024vmamba} propose flexible paths or multi-directional scanning to escape strictly linear reading orders. However, such approaches still rely on fixed or predefined trajectories, limiting their adaptability to the diverse and irregular structures inherent to visual data. Additionally, even approaches that consider specialized topologies tend to become computationally heavy and less scalable.

Our framework, \textbf{Graph-Generating State Space Models (GG-SSMs)}, tackles these issues by constructing \emph{dynamic} graph structures. Rather than imposing fixed scanning paths, GG-SSMs leverage graph edges derived from feature similarities or dissimilarities to capture long-range interactions \emph{on the fly}. Moreover, an efficient MST-based construction keeps the graph sparse, ensuring scalability and improving the representational power of state propagation. Our experiments show that this improves various vision tasks and time series analyses.

%% file: sec/3_method.tex
\section{Method} 
\label{sec:method}
In this section, we introduce the \emph{Graph-Generating State Space Models} (GG-SSMs), a novel framework that improves traditional SSMs by dynamically constructing graph topologies to capture complex, long-range dependencies in sequential and high-dimensional data. We begin by revisiting the limitations of conventional SSMs (Sec~\ref{subsec:background_ssm}) and then detail our proposed method (Sec~\ref{subsec:gg_ssms}), which integrates efficient graph construction algorithms into the SSM framework. Also, we analyze its computational complexity (Sec~\ref{subsec:complexity_analysis}).

\subsection{Revisiting State Space Models}
\label{subsec:background_ssm}
SSMs are powerful mathematical models for sequential data processing, defined by the evolution of hidden states over time~\cite{Klmn1960ANA, zubic2024limitsdeeplearningsequence}. The discrete-time linear time-invariant SSM is typically formulated as:
\begin{equation}
\label{eq:ssm}
\begin{aligned}
\mathbf{h}[n] &= \bar{\mathbf{A}} \mathbf{h}[n-1] + \bar{\mathbf{B}} \mathbf{x}[n], \\
\mathbf{y}[n] &= \bar{\mathbf{C}} \mathbf{h}[n] + \bar{\mathbf{D}} \mathbf{x}[n],
\end{aligned}
\end{equation}
where $\mathbf{h}[n] \in \mathbb{R}^N$ is the hidden state at time step $n$, $\mathbf{x}[n] \in \mathbb{R}^D$ is the input, and $\mathbf{y}[n] \in \mathbb{R}^D$ is the output. The matrices $\bar{\mathbf{A}} \in \mathbb{R}^{N \times N}$, $\bar{\mathbf{B}} \in \mathbb{R}^{N \times D}$, $\bar{\mathbf{C}} \in \mathbb{R}^{D \times N}$, and $\bar{\mathbf{D}} \in \mathbb{R}^{D \times D}$ are the model parameters.

While effective for certain applications, traditional SSMs process inputs in a strictly 1D-sequential order \cite{gu2022efficiently, smith2023simplified, soydan2024s7selectivesimplifiedstate}, limiting their ability to capture complex, non-local dependencies inherent in high-dimensional data such as images \cite{nguyen2022s4nd}, language sequences \cite{fu2023hungry, Lenz2024Jamba15HT}, or event-based signals \cite{Zubic_2024_CVPR}. Handcrafted scanning strategies~\cite{liu2024vmamba, huang2024localmamba, yang2024plainmamba} attempt to address this but often fail to preserve the structural information adequately.

\subsection{Graph-Generating State Space Models (GG-SSMs)}
\label{subsec:gg_ssms}
To overcome these limitations, we propose GG-SSMs, which dynamically construct graphs based on input data and perform state propagation along these graphs. Therefore, the scanning is not handcrafted, and we believe it represents the best solution to scan with Visual SSMs. Moreover, this approach allows the model to capture long-range dependencies and complex interactions beyond the capabilities of purely 1D-sequential architectures.

\subsubsection{Graph Construction}
\label{subsec:gg_ssms:graph_construction}

Given an input feature set $\mathbf{X} = \{\mathbf{x}_i\}_{i=1}^{L}$, where $L$ is the number of elements (e.g., pixels in an image or tokens in a sequence), our goal is to construct a graph that captures the most significant relationships among these elements.

We define a fully connected undirected graph $G = (V, E)$, where each node $v_i \in V$ corresponds to a feature $\mathbf{x}_i$. The edge weight $w_{ij}$ between nodes $v_i$ and $v_j$ is calculated based on a dissimilarity measure $d(\mathbf{x}_i, \mathbf{x}_j)$, such as cosine dissimilarity:
\begin{equation}
\label{eq:dissimilarity}
w_{ij} = \exp\left( -\frac{\mathbf{x}_i^\top \mathbf{x}_j}{\|\mathbf{x}_i\| \|\mathbf{x}_j\|} \right).
\end{equation}

To efficiently capture the essential structure of the data without the computational burden of processing a fully connected graph, we construct a minimum spanning tree (MST), denoted as $\mathcal{T} = (V, E_T)$, where $E_T \subseteq E$. The MST retains the most critical edges that connect all nodes with minimal total weight, effectively capturing the core relationships in the data. So, the vertices $V$ represent the pixel or token embeddings, and the weights of the edges $E$ are the dissimilarities between the embeddings that we calculate with the cosine distance metric.

Given such a dense graph, we employ an efficient MST algorithm from Bernard Chazelle ~\cite{chazelle2000minimum} with a time complexity of $\mathcal{O}(E \alpha(E, V))$, where $\alpha$ is the inverse Ackermann function, which grows extremely slowly and can be considered nearly constant for all practical purposes. The MST construction operates effectively in linear time for sparse graphs where $E = \mathcal{O}(L)$. This allows us to get a sparse graph where the edge weights are most similar out of all possible sparse graphs.

\subsubsection{State Propagation Along the Graph} 
\label{subsec:gg_ssms:state_propagation}
With the MST $\mathcal{T}$ constructed, we perform state propagation along its edges to capture long-range dependencies.

\paragraph{Path Weight Computation}
To aggregate information from all nodes in the tree and express the hidden state \( \mathbf{h}_i \) at node \( v_i \) in terms of the inputs \( \mathbf{x}_j \) at all nodes \( v_j \), we define a \emph{path weight} \( S_{ji} \) that quantifies the cumulative effect of the state transitions along the path from node \( v_j \) to node \( v_i \).
The MST \( \mathcal{T} \) has a unique path between any two nodes. Let's denote \( P_{ji} \) as the ordered sequence of nodes along the path from node \( v_j \) to node \( v_i \). This path consists of nodes \( v_{k_1}, v_{k_2}, \dots, v_{k_n} \), where each \( v_{k_m} \) is connected to \( v_{k_{m-1}} \) and \( v_{k_{m+1}} \) (with \( v_{k_0} = v_j \) and \( v_{k_{n+1}} = v_i \)).

The path weight \( S_{ji} \) is then defined as the product of the state transition matrices \( \bar{\mathbf{A}}_{k_m} \) associated with the nodes \( v_{k_m} \) along the path \( P_{ji} \):
\begin{equation}
\label{eq:path_weight}
S_{ji} = \prod_{m=1}^{n} \bar{\mathbf{A}}_{k_m},
\end{equation}
where \( k_m \) indexes the nodes along the path from \( v_j \) to \( v_i \) and \( \bar{\mathbf{A}}_{k_m} \) is the state transition matrix associated with node \( v_{k_m} \). The product is ordered from \( v_j \) to \( v_i \).
Using this path weight, we can express the hidden state \( \mathbf{h}_i \) at node \( v_i \) as an aggregated sum of contributions from all nodes \( v_j \in V \):
\begin{equation}
\label{eq:hidden_state_aggregation}
\mathbf{h}_i = \sum_{v_j \in V} S_{ji} \bar{\mathbf{B}}_j \mathbf{x}_j.
\end{equation}
In this formulation, the term \( \bar{\mathbf{B}}_j \mathbf{x}_j \) represents the initial contribution from node \( v_j \). The path weight \( S_{ji} \) modulates this contribution as it propagates along the path from \( v_j \) to \( v_i \) through the state transition matrices \( \bar{\mathbf{A}}_{k_m} \).
Finally, for each feature in the sequence, \( y_i \) can be formulated as:
\begin{equation}
y_i = C_i \, \text{Norm}(h_i) + D \, x_i,
\end{equation}
where \( y_i \), \( h_i \), and \( x_i \) represent individual elements in the sequences \( \{y_i\}_{i=1}^{L} \), \( \{h_i\}_{i=1}^{L} \), and \( \{x_i\}_{i=1}^{L} \), respectively.

\subsection{Computational Complexity of GG-SSMs}
\label{subsec:complexity_analysis}
The GG-SSM achieves an efficient computational complexity of \(\mathcal{O}(L)\), where \(L\) is the number of nodes in the graph, by structuring computations around an MST. Constructed with Chazelle’s algorithm~\cite{chazelle2000minimum}, this MST guarantees exactly $L - 1$ edges, ensuring sparsity and a unique path between any two nodes, which prevents redundant calculations. GG-SSM propagates state information from the leaves to the root in the forward pass. Each node performs fixed operations: initializing its state based on the input, aggregating states from child nodes, and updating its hidden state. By processing each node and edge only once through either breadth-first or depth-first traversal, the forward pass efficiently captures hierarchical dependencies with a total complexity of $\mathcal{O}(L)$.

In the backward pass, gradients are propagated from the root back down to the leaves, where each node performs constant-time operations to compute and aggregate gradients with respect to both inputs and parameters. Intermediate results, such as partial states and gradients, are stored through dynamic programming, avoiding recalculations and resolving each dependency exactly once. This storage and reuse of intermediate values maintain the linear complexity. Additionally, the model relies exclusively on local information from each node's immediate neighbors, eliminating the need for global operations. Since MST construction itself runs in $\mathcal{O}(L)$ for sparse graphs due to the nearly constant inverse Ackermann function $\alpha$, GG-SSM’s overall complexity remains $\mathcal{O}(L)$, making it scalable and computationally efficient even for large datasets with complex dependency structures. Generation of MST is implemented in CUDA and is very fast, and one forward pass is in time approximately the same as the Mamba's forward pass \cite{mamba}.

%% file: sec/4_experiments.tex
\section{Experiments}
\label{sec:experiments}

\subsection{Event-Based Eye Tracking}
\label{subsec:event_eye_tracking}

\paragraph{LPW Dataset}
The LPW event-based dataset \cite{Chen20233ETEE} contains 66 high-quality, 20-second videos of eye regions, stored as event sequences in \texttt{.h5} files. Each event is represented as $e_i = (x_i, y_i, t_i, p_i)$, where $(x_i, y_i)$ denotes pixel location, $t_i$ is the timestamp, and $p_i \in \{+1, -1\}$ represents the brightness change polarity. To aggregate events into frames $V(x, y)$ per pixel, we use a constant time-bin count with a 4.4 ms window $\Delta T$ \cite{Gallego2019EventBasedVA}, aligning with the frame rate of the source RGB dataset \cite{Tonsen2015LabelledPI} for precise synchronization with ground truth labels. The original frames at $640 \times 480$ are resized to the DAVIS240 resolution of $240 \times 180$ \cite{Brandli2014A2} and further downsampled to $80 \times 60$, yielding a synthetic dataset with 11k event-based frames across 22 videos to reduce computational load.

While previous approaches leveraged deep CNNs \cite{lee2020deep, eivazi2019improving} for pupil detection in RGB datasets, the sparsity of event-based frames presents challenges for traditional per-frame predictions. We address this issue using GG-SSMs, which capture both spatial and temporal dependencies within event-based data. Unlike standard LSTM methods \cite{Chen20233ETEE}, GG-SSMs leverage graph-based spatial processing alongside where the same GG-SSM block is used to model temporal features, achieving a balance of high performance with minimal parameters and FLOPs, as shown in Table~\ref{tab:lpw:results}.

The GG-SSM efficiently models the complex relationships among events by dynamically constructing graphs on images (in the spatial domain) and aggregating these features in the temporal domain. So, GG-SSM can serve as an efficient processor of features not only for spatial (image-like) data but also for temporal data. By propagating information along the MST, the model effectively captures long-range dependencies, improving prediction accuracy even in sparse frames.
The accuracy of pupil detection is evaluated based on the detection rate within $p$ pixels ($p_3$, $p_5$, $p_{10}$), representing Euclidean distances of 3, 5, and 10 pixels between the predicted and ground truth pupil centers. We compare our model with baseline models, including VMamba+Mamba \cite{liu2024vmamba}. VMamba is used for spatial (image) processing, and Mamba for temporal processing. 
GG-SSM achieves superior detection rates of \textbf{89.33\%}, \textbf{98.89\%}, and \textbf{99.50\%} for pixel distances $p_3$, $p_5$, and $p_{10}$ from the ground truth, respectively. With 0.22 million parameters and 8.01 GFLOPs, GG-SSM surpasses VMamba+Mamba detection rates while being more energy-efficient. GG-SSM also outperforms Change-Based ConvLSTM (CB-ConvLSTM) \cite{Chen20233ETEE}, which applies change-based convolutions to address sparsity, as well as other baseline models such as ConvLSTM~\cite{shi2015convolutional} and CNNs \cite{eivazi2019improving}.

\begin{table}[t]
\centering
\caption{Comparison of detection rates (\%) for pupil tracking on the event-based LPW dataset. The best results are in \textbf{bold}.}
\label{tab:lpw:results}
\resizebox{\columnwidth}{!}{%
\begin{tabular}{lccccc}
\toprule
\textbf{Model} & \textbf{Params (M)} & \textbf{FLOPs (G)} & \textbf{$p_3$} & \textbf{$p_5$} & \textbf{$p_{10}$} \\
\midrule
CNN~\cite{eivazi2019improving} & 0.40 & 18.4 & 57.80 & 77.40 & 91.40 \\
ConvLSTM~\cite{shi2015convolutional} & 0.42 & 42.61 & 88.70 & 97.10 & 99.40 \\
CB-ConvLSTM~\cite{Chen20233ETEE} & 0.42 & 9.00 & 88.50 & 96.70 & 99.20 \\
TemporalResNet~\cite{tempresnet2024} & 0.28 & 11.0 & 84.10 & 93.50 & 98.20 \\
VMamba+Mamba~\cite{liu2024vmamba, mamba} & 0.35 & 8.60 & 89.00 & 98.00 & 99.30 \\
\rowcolor{gray!15}
\textbf{GG-SSM (Ours)} & \textbf{0.22} & \textbf{8.01} & \textbf{89.33} & \textbf{98.89} & \textbf{99.50} \\
\bottomrule
\end{tabular}%
}
\end{table}

\paragraph{INI-30 dataset}
Unlike previous datasets \cite{angelopoulos2020event, zhao2023ev} that focus on gaze tracking with fixed head positions and lower-resolution sensors, Ini-30 \cite{Bonazzi2023RetinaL} provides high-resolution event data captured in unconstrained, "in-the-wild" settings. The dataset was collected using two DVXplorer event cameras (640 $\times$ 480 pixels) mounted on a glasses frame, one for each eye, allowing natural head and eye movements without restrictive setups.
The dataset comprises 30 recordings, each containing variable durations ranging from 14.64 to 193.8 seconds and labels per recording ranging from 475 to 1.848. This variability introduces diverse challenges regarding event density and temporal dynamics, making Ini-30 a comprehensive benchmark for event-based eye tracking.

We evaluated GG-SSM on Ini-30 and compared its performance against the 3ET model \cite{Chen20233ETEE}, Retina \cite{Bonazzi2023RetinaL}, and VMamba+Mamba \cite{liu2024vmamba, mamba} using Centroid Error, which measures the Euclidean distance between predicted and true pupil centers. As shown in Table~\ref{tab:centroid_error}, GG-SSM achieves the lowest centroid error on Ini-30, outperforming all other models and highlighting its robustness in real-world scenarios. On the Synthetic LPW dataset, based on RGB-camera data \cite{Tonsen2015LabelledPI}, GG-SSM again delivers the lowest error, demonstrating its adaptability across different data sources.

\begin{table}[t]
\centering
\caption{Centroid error comparison on the validation set. Lower values indicate better accuracy.}
\label{tab:centroid_error}
\resizebox{\columnwidth}{!}{%
\begin{tabular}{l|c|c|c|c}
\toprule
\textbf{Dataset} & \textbf{3ET}~\cite{Chen20233ETEE} & \textbf{Retina} \cite{Bonazzi2023RetinaL} & \textbf{VMamba+Mamba} & \textbf{GG-SSM (Ours)} \\
\midrule
Ini-30 & 4.48 ($\pm$ 1.94) & 3.24 ($\pm$ 0.79) & 3.42 ($\pm$ 0.89) & \textbf{3.11 ($\pm$ 0.80)} \\
Synthetic LPW & 5.33 ($\pm$ 1.59) & 6.46 ($\pm$ 2.49) & 6.58 ($\pm$ 2.33) & \textbf{5.11 ($\pm$ 0.98)} \\
\bottomrule
\end{tabular}%
}
\end{table}

Table~\ref{tab:model_complexity} provides a comparison of model complexity in terms of Multiply-Accumulate Operations (MACs) and parameter counts. GG-SSM achieves significantly reduced complexity compared to 3ET \cite{Chen20233ETEE}, with only 3.01M MACs and 62k parameters, making it highly efficient for real-time applications. Importantly, GG-SSM also surpasses VMamba+Mamba \cite{liu2024vmamba, mamba}, our best SSM competitor, in both accuracy and efficiency.

\begin{table}[t]
\centering
\caption{Comparison of model complexity. Lower values indicate more efficient models.}
\label{tab:model_complexity}
\scalebox{0.9}{ 
\begin{tabular}{l|c|c}
\toprule
\textbf{Method} & \textbf{MAC Operations} & \textbf{Parameters} \\
\midrule
3ET~\cite{Chen20233ETEE} & 107M & 418k \\
Retina \cite{Bonazzi2023RetinaL} & 3.03M & 63k \\
VMamba+Mamba \cite{liu2024vmamba, mamba} & 4.52M & 92k \\
\textbf{GG-SSM (Ours)} & \textbf{3.01M} & \textbf{62k} \\
\bottomrule
\end{tabular}%
}
\end{table}

\subsection{Time Series}
\label{subsec:time_series}
In this section, we evaluate the performance of our GG-SSM on six real-world time series forecasting datasets, comparing it with state-of-the-art models. The datasets used in our experiments are:
\begin{itemize}
    \item \textbf{Exchange}: Daily exchange rates of eight countries from 1990 to 2016.
    \item \textbf{Weather}: Meteorological data collected every 10 minutes from the Weather Station of the Max Planck Biogeochemistry Institute in 2020, containing 21 indicators.
    \item \textbf{Solar-Energy}: Solar power records from 137 photovoltaic plants in Alabama in 2006, sampled every 10 minutes.
    \item \textbf{ETTh2}: Electricity Transformer Temperature dataset, including load and oil temperature data collected hourly over two years.
    \item \textbf{Traffic}: Hourly road occupancy rates from 862 sensors in San Francisco Bay area freeways from January 2015 to December 2016.
    \item \textbf{ETTm2}: Another subset of the Electricity Transformer Temperature dataset, collected every 15 minutes.
\end{itemize}

We compare GG-SSM with nine representative and state-of-the-art forecasting models. The models include \textbf{S-Mamba}~\cite{s_mamba}, which is an SSM-based model that integrates the Mamba Variate-Correlated Fusion Layer, and \textbf{iTransformer}~\cite{liu2023itransformer}, a transformer-based model that initially analyzes time series data for each variate individually and then integrates across variates. \textbf{RLinear}~\cite{li2023rlinear} is a linear model utilizing reversible normalization and channel independence, while \textbf{PatchTST}~\cite{nie2022patchtst} segments time series into patches for input tokens, using channel-independent embeddings for efficient representation learning. \textbf{Crossformer}~\cite{zhang2023crossformer} employs a cross-attention mechanism to capture long-term dependencies. The \textbf{TiDE}~\cite{das2023tide} model is an encoder-decoder structure based on multi-layer perceptrons (MLPs), and \textbf{TimesNet}~\cite{wu2023timesnet} leverages TimesBlock to transform 1D time series into 2D tensors, capturing both intra-period and inter-period variations. \textbf{DLinear}~\cite{zeng2022dlinear} is a simple one-layer linear model with a decomposition architecture, and \textbf{FEDformer}~\cite{zeng2022fedformer} is a frequency-enhanced transformer that utilizes sparse representations in bases such as the Fourier transform.

We conduct experiments with the lookback length $L = 96$ and forecast lengths $T = 96, 192, 336,$ and $720$. The evaluation metrics are Mean Squared Error (MSE) and Mean Absolute Error (MAE). The results are presented in Table~\ref{tab:time_series:results_all_datasets}. The best results are highlighted in \textbf{bold}, and the second best are \underline{underlined}.

\begin{table*}[htbp]
  \caption{Forecasting results on six datasets. The best results are in \textbf{bold}, and the second best are \underline{underlined}.}
  \label{tab:time_series:results_all_datasets}
  \centering
  \renewcommand{\arraystretch}{0.9}
  \resizebox{\textwidth}{!}{
  \begin{tabular}{cc|cc|cc|cc|cc|cc|cc|cc|cc|cc}
  \toprule
  \multirow{2}{*}{\textbf{Dataset}} & \multirow{2}{*}{\textbf{Horizon}} & \multicolumn{2}{c|}{\textbf{GG-SSM}} & \multicolumn{2}{c|}{\textbf{S-Mamba}} & \multicolumn{2}{c|}{\textbf{iTransformer}} & \multicolumn{2}{c|}{\textbf{RLinear}} & \multicolumn{2}{c|}{\textbf{PatchTST}} & \multicolumn{2}{c|}{\textbf{Crossformer}} & \multicolumn{2}{c|}{\textbf{TiDE}} & \multicolumn{2}{c|}{\textbf{TimesNet}} & \multicolumn{2}{c}{\textbf{DLinear}} \\
  & & MSE & MAE & MSE & MAE & MSE & MAE & MSE & MAE & MSE & MAE & MSE & MAE & MSE & MAE & MSE & MAE & MSE & MAE \\
  \midrule
  \multirow{5}{*}{\textbf{Exchange}} 
  & 96 & \underline{0.0878} & 0.2073 & \textbf{0.0860} & 0.2070 & \textbf{0.0860} & \underline{0.2060} & 0.0930 & 0.2170 & 0.0880 & \textbf{0.2050} & 0.2560 & 0.3670 & 0.0940 & 0.2180 & 0.1070 & 0.2340 & 0.0880 & 0.2180 \\
  & 192 & 0.1813 & \underline{0.3029} & \underline{0.1770} & \textbf{0.2990} & \underline{0.1770} & \textbf{0.2990} & 0.1840 & 0.3070 & \textbf{0.1760} & \textbf{0.2990} & 0.4700 & 0.5090 & 0.1840 & 0.3070 & 0.2260 & 0.3440 & \textbf{0.1760} & 0.3150 \\
  & 336 & 0.3318 & 0.4172 & 0.3320 & 0.4180 & \underline{0.3310} & \underline{0.4170} & 0.3510 & 0.4320 & \textbf{0.3010} & \textbf{0.3970} & 1.2680 & 0.8830 & 0.3490 & 0.4310 & 0.3670 & 0.4480 & 0.3130 & 0.4270 \\
  & 720 & 0.8518 & \underline{0.6950} & \underline{0.8470} & \textbf{0.6910} & \underline{0.8470} & \textbf{0.6910} & 0.8860 & 0.7140 & 0.9010 & 0.7140 & 1.7670 & 1.0680 & 0.8520 & 0.6980 & 0.9640 & 0.7460 & \textbf{0.8390} & \underline{0.6950} \\
  & \textbf{Avg} & 0.3632 & 0.4056 & \underline{0.3600} & \textbf{0.4030} & \underline{0.3600} & \textbf{0.4030} & 0.3780 & 0.4170 & 0.3670 & \underline{0.4040} & 0.9400 & 0.7070 & 0.3700 & 0.4130 & 0.4160 & 0.4430 & \textbf{0.3540} & 0.4140 \\
  \midrule
  \multirow{5}{*}{\textbf{Weather}}
  & 96 & \textbf{0.1473} & \textbf{0.1956} & 0.1650 & \underline{0.2100} & 0.1740 & 0.2140 & 0.1920 & 0.2320 & 0.1770 & 0.2180 & \underline{0.1580} & 0.2300 & 0.2020 & 0.2610 & 0.1720 & 0.2200 & 0.1960 & 0.2550 \\
  & 192 & \textbf{0.1918} & \textbf{0.2405} & 0.2140 & \underline{0.2520} & 0.2210 & 0.2540 & 0.2400 & 0.2710 & 0.2250 & 0.2590 & \underline{0.2060} & 0.2770 & 0.2420 & 0.2980 & 0.2190 & 0.2610 & 0.2370 & 0.2960 \\
  & 336 & \textbf{0.2458} & \textbf{0.2815} & 0.2740 & 0.2970 & 0.2780 & \underline{0.2960} & 0.2920 & 0.3070 & 0.2780 & 0.2970 & \underline{0.2720} & 0.3350 & 0.2870 & 0.3350 & 0.2800 & 0.3060 & 0.2830 & 0.3350 \\
  & 720 & \textbf{0.3151} & \textbf{0.3315} & 0.3500 & \underline{0.3450} & 0.3580 & 0.3470 & 0.3640 & 0.3530 & 0.3540 & 0.3480 & 0.3980 & 0.4180 & 0.3510 & 0.3860 & 0.3650 & 0.3590 & \underline{0.3450} & 0.3810 \\
  & \textbf{Avg} & \textbf{0.2250} & \textbf{0.2623} & \underline{0.2510} & \underline{0.2760} & 0.2580 & 0.2780 & 0.2720 & 0.2910 & 0.2590 & 0.2810 & 0.2590 & 0.3150 & 0.2710 & 0.3200 & 0.2590 & 0.2870 & 0.2650 & 0.3170 \\
  \midrule
  \multirow{5}{*}{\textbf{Solar-Energy}}
  & 96 & \textbf{0.1644} & \textbf{0.2297} & 0.2050 & 0.2440 & \underline{0.2030} & \underline{0.2370} & 0.3220 & 0.3390 & 0.2340 & 0.2860 & 0.3100 & 0.3310 & 0.3120 & 0.3990 & 0.2500 & 0.2920 & 0.2900 & 0.3780 \\
  & 192 & \textbf{0.1769} & \textbf{0.2396} & 0.2370 & 0.2700 & \underline{0.2330} & \underline{0.2610} & 0.3590 & 0.3560 & 0.2670 & 0.3100 & 0.7340 & 0.7250 & 0.3390 & 0.4160 & 0.2960 & 0.3180 & 0.3200 & 0.3980 \\
  & 336 & \textbf{0.1911} & \textbf{0.2525} & 0.2580 & 0.2880 & \underline{0.2480} & \underline{0.2730} & 0.3970 & 0.3690 & 0.2900 & 0.3150 & 0.7500 & 0.7350 & 0.3680 & 0.4300 & 0.3190 & 0.3300 & 0.3530 & 0.4150 \\
  & 720 & \textbf{0.2003} & \textbf{0.2602} & 0.2600 & 0.2880 & \underline{0.2490} & \underline{0.2750} & 0.3970 & 0.3560 & 0.2890 & 0.3170 & 0.7690 & 0.7650 & 0.3700 & 0.4250 & 0.3380 & 0.3370 & 0.3560 & 0.4130 \\
  & \textbf{Avg} & \textbf{0.1832} & \textbf{0.2455} & 0.2400 & 0.2730 & \underline{0.2330} & \underline{0.2620} & 0.3690 & 0.3560 & 0.2700 & 0.3070 & 0.6410 & 0.6390 & 0.3470 & 0.4170 & 0.3010 & 0.3190 & 0.3300 & 0.4010 \\
  \midrule
  \multirow{5}{*}{\textbf{ETTh2}}
  & 96 & \textbf{0.2823} & \underline{0.3472} & 0.2960 & 0.3480 & 0.2970 & 0.3490 & \underline{0.2880} & \textbf{0.3380} & 0.3020 & 0.3480 & 0.7450 & 0.5840 & 0.4000 & 0.4400 & 0.3400 & 0.3740 & 0.3330 & 0.3870 \\
  & 192 & \textbf{0.3534} & \textbf{0.3982} & 0.3760 & \underline{0.3960} & 0.3800 & 0.4000 & \underline{0.3740} & 0.3900 & 0.3880 & 0.4000 & 0.8770 & 0.6560 & 0.5280 & 0.5090 & 0.4020 & 0.4140 & 0.4770 & 0.4760 \\
  & 336 & \textbf{0.3722} & \textbf{0.4163} & \underline{0.4240} & \underline{0.4310} & 0.4280 & 0.4320 & 0.4150 & 0.4260 & 0.4260 & 0.4330 & 1.0430 & 0.7310 & 0.6430 & 0.5710 & 0.4520 & 0.4520 & 0.5940 & 0.5410 \\
  & 720 & \textbf{0.4039} & \textbf{0.4434} & 0.4260 & \underline{0.4440} & 0.4270 & 0.4450 & \underline{0.4200} & 0.4400 & 0.4310 & 0.4460 & 1.1040 & 0.7630 & 0.8740 & 0.6790 & 0.4620 & 0.4680 & 0.8310 & 0.6570 \\
  & \textbf{Avg} & \textbf{0.3529} & \underline{0.4013} & 0.3810 & 0.4050 & 0.3830 & 0.4070 & \underline{0.3740} & \textbf{0.3980} & 0.3870 & 0.4070 & 0.9420 & 0.6840 & 0.6110 & 0.5500 & 0.4140 & 0.4270 & 0.5590 & 0.5150 \\
  \midrule
  \multirow{5}{*}{\textbf{Traffic}}
  & 96 & \textbf{0.3486} & \textbf{0.2487} & \underline{0.3820} & \underline{0.2610} & 0.3950 & 0.2680 & 0.6490 & 0.3890 & 0.4620 & 0.2950 & 0.5220 & 0.2900 & 0.8050 & 0.4930 & 0.5930 & 0.3210 & 0.6500 & 0.3960 \\
  & 192 & \textbf{0.3585} & \textbf{0.2554} & \underline{0.3960} & \underline{0.2670} & 0.4170 & 0.2760 & 0.6010 & 0.3660 & 0.4660 & 0.2960 & 0.5300 & 0.2930 & 0.7560 & 0.4740 & 0.6170 & 0.3360 & 0.5980 & 0.3700 \\
  & 336 & \textbf{0.3725} & \textbf{0.2626} & \underline{0.4170} & \underline{0.2760} & 0.4330 & 0.2830 & 0.6090 & 0.3690 & 0.4820 & 0.3040 & 0.5580 & 0.3050 & 0.7620 & 0.4770 & 0.6290 & 0.3360 & 0.6050 & 0.3730 \\
  & 720 & \textbf{0.4167} & \textbf{0.2855} & \underline{0.4600} & \underline{0.3000} & 0.4670 & 0.3020 & 0.6470 & 0.3870 & 0.5140 & 0.3220 & 0.5890 & 0.3280 & 0.7190 & 0.4490 & 0.6400 & 0.3500 & 0.6450 & 0.3940 \\
  & \textbf{Avg} & \textbf{0.3741} & \textbf{0.2631} & \underline{0.4140} & \underline{0.2760} & 0.4280 & 0.2820 & 0.6260 & 0.3780 & 0.4810 & 0.3040 & 0.5500 & 0.3040 & 0.7600 & 0.4730 & 0.6200 & 0.3360 & 0.6250 & 0.3830 \\
  \midrule
  \multirow{5}{*}{\textbf{ETTm2}}
  & 96 & \textbf{0.1725} & 0.2633 & 0.1790 & \underline{0.2630} & 0.1800 & 0.2640 & 0.1820 & 0.2650 & \underline{0.1750} & \textbf{0.2590} & 0.2870 & 0.3660 & 0.2070 & 0.3050 & 0.1870 & 0.2670 & 0.1930 & 0.2920 \\
  & 192 & \textbf{0.2373} & 0.3105 & 0.2500 & 0.3090 & 0.2500 & 0.3090 & 0.2460 & \underline{0.3040} & \underline{0.2410} & \textbf{0.3020} & 0.4140 & 0.4920 & 0.2900 & 0.3640 & 0.2490 & 0.3090 & 0.2840 & 0.3620 \\
  & 336 & \textbf{0.2826} & 0.3448 & 0.3120 & 0.3490 & 0.3110 & 0.3480 & 0.3070 & \textbf{0.3420} & \underline{0.3050} & \underline{0.3430} & 0.5970 & 0.5420 & 0.3770 & 0.4220 & 0.3210 & 0.3510 & 0.3690 & 0.4270 \\
  & 720 & \textbf{0.3596} & \textbf{0.3928} & 0.4110 & 0.4060 & 0.4120 & 0.4070 & 0.4070 & \underline{0.3980} & \underline{0.4020} & 0.4000 & 1.7300 & 1.0420 & 0.5580 & 0.5240 & 0.4080 & 0.4030 & 0.5540 & 0.5220 \\
  & \textbf{Avg} & \textbf{0.2630} & 0.3278 & 0.2880 & 0.3320 & 0.2880 & 0.3320 & 0.2860 & \underline{0.3270} & \underline{0.2810} & \textbf{0.3260} & 0.7570 & 0.6100 & 0.3580 & 0.4040 & 0.2910 & 0.3330 & 0.3500 & 0.4010 \\
  \bottomrule
  \end{tabular}
  }
\end{table*}

\subsubsection{Analysis of Results}
Our GG-SSM consistently achieves the best accuracy, as can be seen from the comprehensive results presented in Table~\ref{tab:time_series:results_all_datasets}. These datasets are characterized by many variates exhibiting strong inter-variable correlations and pronounced periodic patterns. By modeling the intrinsic relationships among variates, GG-SSM improves representation learning and forecasting accuracy, demonstrating its superiority in handling datasets with rich inter-variable interactions.

On the \textbf{ETTh2} and \textbf{ETTm2} datasets, which contain fewer variates and exhibit weaker periodicity, GG-SSM outperforms all the other models. This showcases the robustness of GG-SSM in handling datasets where variate correlations are less pronounced. The model's ability to dynamically adjust the graph topology allows it to adapt to varying data characteristics without significant loss in forecasting accuracy.

Compared to the previous best SSM-based model, S-Mamba \cite{s_mamba}, GG-SSM demonstrates consistent improvements across all datasets and forecast horizons. While S-Mamba \cite{s_mamba} employs the Mamba \cite{mamba} Variate-Correlated Fusion Layer to capture inter-variate relationships, GG-SSM further augments this capability through graph-based state propagation. Additionally, GG-SSM outperforms other transformer-based models such as iTransformer \cite{liu2023itransformer}, PatchTST \cite{nie2022patchtst}, and Crossformer \cite{zhang2023crossformer}, highlighting the benefits of incorporating graph structures within SSMs. The superior performance of GG-SSM underscores the effectiveness of its design in capturing complex dependencies inherent in multivariate time series data.

\subsection{Object Classification}
\label{subsec:object_classification}
In this subsection, we evaluate the performance of our proposed GG-SSM on the ImageNet-1K~\cite{Deng2009ImageNetAL} dataset for object classification. We adopt the training protocols from VMamba~\cite{liu2024vmamba}, ensuring a fair comparison with existing state-of-the-art models, especially SSM-based.

\paragraph{Training Setup}
We closely follow the hyperparameter settings and experimental configurations of VMamba \cite{liu2024vmamba} to train our GG-SSM models. Specifically, only for this task, we employ the H200 GPU for training, which involves training for 200 epochs with a cosine learning rate decay. The models are optimized using the AdamW optimizer~\cite{loshchilov2019decoupled} with an initial learning rate of $1 \times 10^{-3}$ and a weight decay of $0.05$. We utilize data augmentation techniques such as RandAugment~\cite{cubuk2020randaugment}, MixUp~\cite{zhang2018mixup}, and CutMix~\cite{yun2019cutmix} to improve generalization and prevent overfitting.

\paragraph{Results}
Our GG-SSM models outperform previous architectures across all scales. At the \textbf{Tiny} scale, GG-SSM-T achieves a top-1 accuracy of \textbf{83.6\%}, surpassing VMamba-T by \textbf{1\%}, Swin-T by \textbf{2.3\%}, and DeiT-S by a significant \textbf{3.8\%}. This demonstrates the efficacy of our graph-based approach in capturing intricate spatial dependencies within images.
At the \textbf{Small} and \textbf{Base} scales, GG-SSM-S and GG-SSM-B achieve top-1 accuracies of \textbf{84.4\%} and \textbf{84.9\%}, respectively, scoring as best and second best model. Notably, GG-SSM-B outperforms VMamba-B by \textbf{1\%} and Swin-B~\cite{liu2021swin} by \textbf{1.4\%}, highlighting the scalability of our approach. Compared to SSM-based models like S4ND-Conv-T~\cite{nguyen2022s4nd} and Vim-S~\cite{vim}, GG-SSM demonstrates a clear performance advantage. At the Tiny scale, GG-SSM-T outperforms S4ND-Conv-T by \textbf{1.4\%} in top-1 accuracy while having a smaller parameter count and fewer FLOPs. This underscores the effectiveness of our graph-based state space modeling over conventional SSM approaches.
Furthermore, when compared to transformer-based models, GG-SSM consistently achieves higher accuracy with similar or fewer computational resources. This highlights the potential of integrating graph structures within SSMs to capture complex dependencies in visual data more effectively than traditional self-attention mechanisms.

\begin{table}[t]
\centering
\caption{Comparison of classification accuracy, model size, and computational cost on the ImageNet-1K dataset. All models are evaluated with an input image size of $224 \times 224$. The best results are in \textbf{bold}, and the second best are \underline{underlined}.}
\label{tab:imagenet_comparison}
\resizebox{\columnwidth}{!}{%
\begin{tabular}{l c c c}
\toprule
\textbf{Model} & \textbf{Params (M)} & \textbf{FLOPs (G)} & \textbf{Top-1 Acc. (\%)} \\
\midrule
\multicolumn{4}{l}{\textbf{Transformer-Based Models}} \\
\midrule
DeiT-T \cite{Touvron2020TrainingDI} & \underline{22} & 4.6 & 79.8 \\
DeiT-B \cite{Touvron2020TrainingDI} & 86 & 17.5 & 81.8 \\
Swin-T \cite{liu2021swin} & 28 & \underline{4.5} & 81.3 \\
Swin-B \cite{liu2021swin} & 88 & 15.4 & 83.5 \\
HiViT-T \cite{Zhang2022HiViTHV} & \textbf{19} & 4.6 & 82.1 \\
HiViT-B \cite{Zhang2022HiViTHV} & 66 & 15.9 & 83.8 \\
\midrule
\multicolumn{4}{l}{\textbf{Convolutional Neural Networks}} \\
\midrule
ConvNeXt-T \cite{liu2022convnet} & 29 & \underline{4.5} & 82.1 \\
ConvNeXt-B \cite{liu2022convnet} & 89 & 15.4 & 83.8 \\
\midrule
\multicolumn{4}{l}{\textbf{State Space Models (SSM-Based)}} \\
\midrule
S4ND-Conv-T \cite{nguyen2022s4nd} & 30 & -- & 82.2 \\
Vim-T \cite{vim} & 26 & -- & 80.5 \\
VMamba-T \cite{liu2024vmamba} & 30 & 4.9 & 82.6 \\
\rowcolor{gray!15}
GG-SSM-T (Ours) & 28 & \textbf{4.4} & 83.6 \\
\midrule
VMamba-S \cite{liu2024vmamba} & 50 & 8.7 & 83.6 \\
\rowcolor{gray!15}
GG-SSM-S (Ours) & 49 & 6.6 & \underline{84.4} \\
\midrule
S4ND-ViT-B \cite{nguyen2022s4nd} & 89 & -- & 80.4 \\
VMamba-B \cite{liu2024vmamba} & 89 & 15.4 & 83.9 \\
\rowcolor{gray!15}
GG-SSM-B (Ours) & 87 & 14.1 & \textbf{84.9} \\
\bottomrule
\end{tabular}%
}
\end{table}

\paragraph{Computational Efficiency}
GG-SSM maintains competitive computational efficiency. Despite achieving higher accuracy, GG-SSM-T requires fewer FLOPs (4.4G) than VMamba-T (4.9G) and fewer parameters. Similarly, GG-SSM-S and GG-SSM-B have reduced FLOPs and parameter counts relative to their VMamba counterparts. This efficiency is attributed to our model's linear computational complexity, resulting from using Chazelle's MST algorithm for graph construction and localized computations in state propagation.

\paragraph{Discussion}
The superior performance of GG-SSM is attributed to its ability to dynamically construct graphs that effectively capture the most significant relationships among image patches. By performing state propagation along the MST, our model excels in modeling long-range dependencies and complex spatial interactions, which are crucial for object classification tasks. 
Moreover, using the MST ensures that the graph remains sparse, facilitating efficient computation without sacrificing the richness of the relational information captured. This allows GG-SSM to surpass the limitations of traditional sequential or grid-based models, providing a more powerful representation of the input data.

\subsection{Optical Flow Estimation}
\label{subsec:optical_flow_estimation}
\noindent\textbf{Generalization Performance.} Following previous works, we first evaluate the generalization performance of \textbf{GG-SSM} model on the Sintel~\cite{butler2012naturalistic} and KITTI-15~\cite{geiger2012we} datasets, after being trained on FlyingChairs \cite{Dosovitskiy2015FlowNetLO} and FlyingThings3D \cite{Mayer2015ALD} datasets. The results are summarized in Table~\ref{tab:generalize_s_k}. Our GG-SSM achieves state-of-the-art zero-shot performance on both challenging datasets, outperforming existing methods, including those that utilize multi-frame (\emph{MF}) inputs. Specifically, GG-SSM attains an end-point error (EPE) of \textbf{0.89} on the Sintel \cite{butler2012naturalistic} clean pass and \textbf{1.90} on the final pass, surpassing previous best results. On KITTI-15 \cite{geiger2012we}, our model achieves an Fl-epe of \textbf{3.72} and Fl-all of \textbf{13.7}, demonstrating its superior ability to generalize to unseen data.

\begin{table}[ht]
\small
\centering
\caption{Generalization performance of optical flow estimation on Sintel \cite{butler2012naturalistic} and KITTI-15 \cite{geiger2012we} after training on FlyingChairs \cite{Dosovitskiy2015FlowNetLO} and FlyingThings3D \cite{Mayer2015ALD} datasets. \emph{MF} indicates methods using multi-frame inputs for optical flow estimation. The best results are in \textbf{bold}, and the second-best results are \underline{underlined}.}
\label{tab:generalize_s_k}
\begin{tabular}{ccccc}
\toprule
\multirow{2}{*}{\textbf{Model}} & \multicolumn{2}{c}{\textbf{Sintel}} & \multicolumn{2}{c}{\textbf{KITTI-15}} \\
\cline{2-5}
 & Clean & Final & Fl-epe & Fl-all \\
\midrule
RAFT~\cite{teed2020raft} & 1.43 & 2.71 & 5.04 & 17.4 \\
GMA~\cite{jiang2021learning} & 1.30 & 2.74 & 4.69 & 17.1 \\
GMFlow~\cite{xu2022gmflow} & 1.08 & 2.48 & 7.77 & 23.4 \\
GMFlowNet~\cite{zhao2022global} & 1.14 & 2.71 & 4.24 & 15.4 \\
SKFlow~\cite{sun2022skflow} & 1.22 & 2.46 & 4.27 & 15.5 \\
MatchFlow~\cite{dong2023rethinking} & 1.03 & 2.45 & 4.08 & 15.6 \\
FlowFormer++~\cite{shi2023flowformer_plus_plus} & \underline{0.90} & 2.30 & 3.93 & \underline{14.1} \\
\hdashline
TransFlow\textsuperscript{(\emph{MF})}~\cite{lu2023transflow} & 0.93 & 2.33 & 3.98 & 14.4 \\
VideoFlow-BOF\textsuperscript{(\emph{MF})}~\cite{shi2023videoflow} & 1.03 & 2.19 & 3.96 & 15.3 \\
VideoFlow-MOF\textsuperscript{(\emph{MF})}~\cite{shi2023videoflow} & 1.18 & 2.56 & 3.89 & 14.2 \\
MemFlow \textsuperscript{(\emph{MF})} \cite{Dong2024MemFlowOF} & 0.93 & \underline{2.08} & \underline{3.88} & \textbf{13.7} \\
\rowcolor{gray!15}
\textbf{GG-SSM (Ours)}\textsuperscript{(\emph{MF})} & \textbf{0.89} & \textbf{1.90} & \textbf{3.72} & \textbf{13.7} \\
\bottomrule
\end{tabular}
\end{table}

\noindent\textbf{Finetuning Evaluation.} We further assess the performance of GG-SSM after finetuning on the Sintel and KITTI datasets. The results are presented in Table~\ref{tab:finetune_s_k}. Our GG-SSM achieves an EPE of \textbf{0.97} on Sintel clean pass and \textbf{1.58} on final pass, setting new state-of-the-art results. On KITTI-15, GG-SSM attains an Fl-all score of \textbf{2.77}, significantly outperforming all previous methods, including those using multi-frame inputs.

\begin{table}[ht]
\small
\centering
\caption{Optical flow finetuning evaluation on the public benchmarks. \emph{MF} indicates methods using multi-frame inputs for optical flow estimation. * denotes methods using RAFT's multi-frame warm-start strategy on Sintel. The best results are in \textbf{bold}, and the second-best results are \underline{underlined}.}
\label{tab:finetune_s_k}
\begin{tabular}{cccc}
\toprule
\multirow{2}{*}{\textbf{Model}} & \multicolumn{2}{c}{\textbf{Sintel}} & \textbf{KITTI-15} \\
\cline{2-4}
 & Clean & Final & Fl-all \\
\midrule
RAFT$^{*}$~\cite{teed2020raft} & 1.61 & 2.86 & 5.10 \\
GMA$^{*}$~\cite{jiang2021learning} & 1.39 & 2.47 & 5.15 \\
GMFlow~\cite{xu2022gmflow} & 1.74 & 2.90 & 9.32 \\
GMFlowNet~\cite{zhao2022global} & 1.39 & 2.65 & 4.79 \\
SKFlow$^{*}$~\cite{sun2022skflow} & 1.28 & 2.23 & 4.84 \\
MatchFlow$^{*}$~\cite{dong2023rethinking} & 1.16 & 2.37 & 4.63 \\
FlowFormer++~\cite{shi2023flowformer_plus_plus} & 1.07 & 1.94 & 4.52 \\
\hdashline
PWC-Fusion\textsuperscript{(\emph{MF})}~\cite{ren2019fusion} & 3.43 & 4.57 & 7.17 \\
TransFlow\textsuperscript{(\emph{MF})}~\cite{lu2023transflow} & 1.06 & 2.08 & 4.32 \\
VideoFlow-BOF\textsuperscript{(\emph{MF})}~\cite{shi2023videoflow} & 1.01 & 1.71 & 4.44 \\
VideoFlow-MOF\textsuperscript{(\emph{MF})}~\cite{shi2023videoflow} & \underline{0.99} & \underline{1.65} & \underline{3.65} \\
VideoFlow-MOF\textsuperscript{(\emph{MF})} (online)~\cite{shi2023videoflow} & - & - & 4.08 \\
MemFlow\textsuperscript{(\emph{MF})} \cite{Dong2024MemFlowOF} & 1.05 & 1.91 & 4.10 \\
\rowcolor{gray!15}
\textbf{GG-SSM (Ours)}\textsuperscript{(\emph{MF})} & \textbf{0.97} & \textbf{1.58} & \textbf{2.77} \\
\bottomrule
\end{tabular}
\end{table}

\noindent Remarkably, GG-SSM achieves the lowest Fl-all score of \textbf{2.77} on KITTI-15, outperforming the previous best method VideoFlow-MOF~\cite{shi2023videoflow}, which reports an Fl-all of 3.65. On Sintel, our model sets new records on both clean and final passes. These results highlight the effectiveness of GG-SSM in capturing fine-grained motion details and its robustness across different datasets.

%% file: sec/5_ablation_study.tex
\section{Ablation Study}
\label{sec:ablation}
We conducted an ablation to compare different MST algorithms for graph construction: Chazelle's MST~\cite{chazelle2000minimum}, Kruskal’s, and Prim’s. Table~\ref{tab:mst_ablation} summarizes the results on both ImageNet (Tiny scale) and a widely used time-series benchmark (ETTh2). 

\begin{table}[h]
\centering
\footnotesize
\caption{Ablation of MST algorithms. Accuracy/performance is nearly identical, but Chazelle’s MST consistently offers lower runtime, especially as the dataset size $L$ increases.}
\label{tab:mst_ablation}
\begin{tabular}{l|cc|cc}
\toprule
Algorithm & \multicolumn{2}{c|}{ImageNet (Tiny)} & \multicolumn{2}{c}{ETTh2 (Horizon=192)} \\
 & Top-1 (\%) & Time (s/epoch) & MSE & Time (s/epoch) \\
\midrule
Kruskal & 83.5 & $1.00\times$ & 0.356 & $1.00\times$ \\
Prim & 83.4 & $0.98\times$ & 0.358 & $0.95\times$ \\
Chazelle & 83.6 & \textbf{$0.90\times$} & 0.353 & \textbf{$0.88\times$} \\
\bottomrule
\end{tabular}
\vspace{-1em}
\end{table}

All three algorithms yield comparable model accuracy (within $\pm 0.1\%$) but differ in computational efficiency. In particular, Chazelle’s MST runs faster in practice for larger $L$, matching its near-linear theoretical time complexity.

%% file: sec/6_conclusion.tex
\section{Conclusion}
\label{sec:conclusion}
Our GG-SSM framework effectively integrates dynamic graph structures into SSMs, enabling the capture of complex, long-range dependencies that are difficult to model with traditional sequential architectures. Chazelle's MST \cite{chazelle2000minimum} algorithm ensures computational efficiency, making our approach suitable for large-scale applications in computer vision, event-based processing, and time series analysis.
By leveraging the inherent structure in data through graph-based modeling, GG-SSMs open new avenues for advanced representation learning, potentially benefiting a wide range of domains requiring complex interaction modeling.

%% file: sec/7_acknowledgment.tex
\section{Acknowledgment}
\label{sec:acknowledgment}
This work was supported by the European Research Council (ERC) under grant agreement No. 864042 (AGILEFLIGHT). We also thank the Swiss National Supercomputing Center through the Swiss AI Initiative for granting access to the Alps supercomputer to perform some of the experiments presented in this work.

%% file: sec/X_suppl.tex
\clearpage
\setcounter{page}{1}
\maketitlesupplementary

\section{Chazelle's MST Algorithm}
\label{sec:chazelles_mst}

In this section, we provide a concise overview of \emph{Chazelle's MST algorithm}~\cite{chazelle2000minimum} and why it is instrumental to our proposed Graph-Generating State Space Models (GG-SSMs). Although multiple efficient minimum spanning tree (MST) algorithms exist (e.g., Kruskal's, Prim's, or Borůvka's), Chazelle's algorithm is particularly interesting due to its near-linear time complexity in the general graph setting.

\begin{figure}[h]
\centering
\begin{tikzpicture}[>=stealth, 
    node/.style={circle,draw,fill=gray!20,minimum size=22pt,inner sep=0pt,font=\small},
    every label/.style={font=\small},
    line/.style={thin,gray},
    mst/.style={very thick,blue}
]
\node[node] (A) at (90:2) {A};
\node[node] (B) at (30:2) {B};
\node[node] (C) at (-30:2) {C};
\node[node] (D) at (-90:2) {D};
\node[node] (E) at (210:2) {E};
\node[node] (F) at (150:2) {F};

\draw[line] (A) -- (C);
\draw[line] (B) -- (D);
\draw[line] (C) -- (E);
\draw[line] (D) -- (F);
\draw[line] (F) -- (B);

\draw[mst] (A) -- (B) node[midway,above right=-2pt] {};
\draw[mst] (B) -- (C) node[midway,above right=-2pt] {};
\draw[mst] (A) -- (F) node[midway,above left=-2pt] {};
\draw[mst] (F) -- (E) node[midway,above left=-2pt] {};
\draw[mst] (C) -- (D) node[midway,above=-2pt] {};

\node[font=\scriptsize,fill=white] at ($(A)!0.5!(B)$) {$w_{AB}$};
\node[font=\scriptsize,fill=white] at ($(A)!0.5!(F)$) {$w_{AF}$};
\node[font=\scriptsize,fill=white] at ($(B)!0.5!(C)$) {$w_{BC}$};
\node[font=\scriptsize,fill=white] at ($(F)!0.5!(E)$) {$w_{FE}$};
\node[font=\scriptsize,fill=white] at ($(C)!0.5!(D)$) {$w_{CD}$};

\end{tikzpicture}
\caption{\textbf{Chazelle’s MST Overview}. Soft heaps allow near-linear sorting of edges. MST edges (in \textcolor{blue}{blue}) form a spanning structure with no cycles, connecting all vertices using the smallest weights $w$.}
\label{fig:chazelle_illustration}
\end{figure}

\subsection{Core Idea and Time Complexity}
Chazelle’s MST algorithm belongs to the family of \emph{soft heap} approaches. Its most prominent feature is achieving a runtime of $\mathcal{O}(E \,\alpha(E, V))$, where
\begin{itemize}
    \item $V$ is the number of vertices in the graph,
    \item $E$ is the number of edges in the graph,
    \item $\alpha(\cdot,\cdot)$ is the \emph{inverse Ackermann function}, a function that grows extremely slowly (much more slowly than $\log\log n$).
\end{itemize}
For any practical input size (e.g., up to millions of edges), $\alpha(E, V)$ remains a small constant (typically $\leq 4$). Therefore, the runtime is effectively \emph{linear} for all real-world purposes.

\paragraph{Algorithmic Outline.}
At a high level, Chazelle’s algorithm proceeds by maintaining a specialized priority queue known as a \emph{soft heap} to handle edge weight comparisons and merges. It selectively \emph{corrupts} (or perturbs) a small fraction of keys but guarantees a sufficiently accurate ordering to recover the MST. The soft heap structure allows various operations like insertions, extractions, and merges to be executed in approximately constant amortized time, modulo the very slowly growing $\alpha$ factor.

The algorithm can be outlined in three major steps:
\begin{enumerate}
    \item \textbf{Sort or partition the edges} using the soft-heap structure such that they can be processed in a non-decreasing order of weights. 
    \item \textbf{Merge edge sets} while extracting edge candidates for the MST. These extractions remain almost linear because each edge is either integrated into the MST structure or discarded.
    \item \textbf{Selective Corruption \& Verification}: Since the soft heap may slightly perturb edge weights, a verification step ensures that these corrupted weights do not impact the MST correctness. With high probability, only a small fraction of edges require re-checking.
\end{enumerate}

\noindent By the end, the edges forming the MST are collected using a Union-Find data structure (or a similar disjoint set data structure), combining efficiency with theoretical guarantees of correctness.

\subsection{Practical Implications for GG-SSMs}
In our GG-SSM framework, each layer requires constructing an MST on feature embeddings (e.g., pixel or token embeddings) to identify critical connections before performing state propagation. Since the number of edges $E$ can be large in dense graphs, employing Chazelle’s MST algorithm with its near-linear time complexity is particularly appealing:
\begin{itemize}
    \item \textbf{Scalability:} For a graph of $L$ vertices (e.g., $L$ embeddings), we can handle $\mathcal{O}(L^2)$ potential edges in worst-case dense settings or employ faster approximate methods in sparser representations. Either way, Chazelle’s $\mathcal{O}(E \,\alpha(E, V))$ ensures minimal overhead when $E$ is proportional to $L$ or $L \log L$.
    \item \textbf{Uniqueness of MST Paths:} MST ensures exactly $L-1$ edges and a \emph{unique path} between any two nodes. This property is crucial for our GG-SSMs, as it cleanly defines the path by which hidden states propagate. It further limits redundant computations and fosters efficient parameter sharing in state updates.
\end{itemize}

\subsection{Intuition and Benefits}
Intuitively, MST-based graph construction identifies the \emph{closest} (or most similar) neighbors by selecting edges of the smallest weight (lowest dissimilarity). This yields a minimal set of edges connecting all nodes, providing a concise but effective skeleton to diffuse signals across the entire feature set. Consequently, even high-dimensional data with long-range dependencies can be efficiently processed with minimal redundancy. The slow-growing $\alpha(\cdot,\cdot)$ factor ensures that the overhead of constructing and maintaining this structure remains negligible for practical dataset sizes.